\documentclass[10pt, a4paper]{article}

\usepackage{lrec-coling2024} % this is the new style

\usepackage{amsmath}
\usepackage{makecell}
\usepackage{CJKutf8}

\title{Audience-specific Explanations for Machine Translation}

\name{Renhan Lou, Jan Niehues} 

\address{Institute for Anthropomatics and Robotics, Karlsruhe Institute of Technology\\
         Karlsruhe, Germany \\
         uqeae@student.kit.edu, jan.niehues@kit.edu}

\abstract{
In machine translation, a common problem is that the translation of certain words even if translated can cause incomprehension of the target language audience due to different cultural backgrounds. A solution to solve this problem is to add explanations for these words. In a first step, we therefore need to identify these words or phrases. In this work we explore techniques to extract example explanations from a parallel corpus. However, the sparsity of sentences containing words that need to be explained makes building the training dataset extremely difficult. In this work, we propose a semi-automatic technique to extract these explanations from a large parallel corpus. Experiments on English$\to$German language pair show that our method is able to extract sentence so that more than 10$\%$ of the sentences contain explanation, while only 1.9$\%$ of the original sentences contain explanations. In addition, experiments on English$\to$French and English$\to$Chinese language pairs also show similar conclusions. This is therefore an essential first automatic step to create a explanation dataset. Furthermore we show that the technique is robust for all three language pairs.
 \\ \newline  }

\begin{document}

\maketitleabstract

\section{Introduction}
With different neural network models proposed, neural machine translation (NMT) is now becoming the dominant approach in machine translation. Especially new neural network models, such as Transformer \citep{vaswani2017attention}, improve significantly the performance of machines in translation tasks. While NMT achieves a very good quality in the direct translation of source text into the target language, in practical application scenarios additional challenges need to be address in order to enable a successful communication across language barriers.

%Although many machine translation models can perform well, they cannot achieve the same high quality as human translation. There are still many problems in translation work that machine translation cannot solve. 

One of the most common problems is that the translation of certain words can cause incomprehension of the audience in the target language. Because some words are common in the source language, but not common in the target language. This leads to the fact that when the audience of the target language sees the translation of these words, they cannot understand the meaning of the translation. 

A simple example is \textbf{\textit{the Super Bowl}}, the annual championship game of the National Football League in the United States. The Super Bowl is one of the most famous games in the United States. However, in some countries in Europe and Asia, only a few people who like American football know about it. Therefore, when the Super Bowl is translated into another language, such as German or Chinese, the audience in the target language will simply understand it as a kind of tableware according to the literal meaning of the translation.

Thus, when these words are translated into another language, how to eliminate the incomprehension of the target language audience is a problem that cannot be ignored in machine translation. Different solutions have been proposed to address the translation difficulties of uncommon words or rare words. The method proposed by \citep{luong2014addressing} uses external dictionaries to overcome the difficulty of rare word translation. Afterwards, \citep{sennrich2015neural} utilize the BPE (byte pair encoding) algorithm to encode rare words into subword units, thereby reducing the number of rare words and improving translation quality. \citep{wu2016google} also use subword units to improve the translation quality of rare words. \citep{pham2018towards} introduce external expert models such as terminology lists to improve the performance of rare word translation.

However, the above work only focuses on the translation of rare words in the source language. Moreover, they cannot eliminate the incomprehension of the target language audience during machine translation. To solve this problem, we can learn from human translation. In human translation, a simple solution to solve this problem is to add explanations when translating these words. With the help of this additional information, the incomprehension of the target language audience can be well eliminated.

%With the help of the human translation solution, the problem of how to eliminate the target language audience’s incomprehension during the translation can be transformed into another more specific problem, that is, How can we model audiences’ specific needs for additional information during translation?

The purpose of our paper is to explore whether it is possible to find a suitable model that can accurately predict which words need to be explained when performing machine translation tasks. In order to train the model the first step is to build a high quality training dataset. However, sentences containing words that need to be explained are extremely uncommon. The sparsity of the target sentence makes it difficult to build the training dataset.

Therefore, we propose a method for finding sentences with words that need to be explained by utilizing external knowledge sources. We conducted experiments on English$\to$German, English$\to$French and English$\to$Chinese language pairs. The results show that the method we propose can greatly reduce the final manual selection work, and at the same time, it can stably and efficiently find the target sentence in the last remaining sentences for all three language pairs. This reduces the difficulty of building a training dataset and also facilitates the training of models in the future. \footnote{Code and data available at:  https://github.com/RHL1014/Audience-specific-Explanations-for-MT}

The contributions of our work are as follows:
\begin{enumerate}
    \item We propose a method for finding sentences with words that need to be explained. Using this method we can find the target sentences stably and efficiently.
    \item This method works for multiple language pairs, and the final effect is independent of the input data distribution. This shows that this method has robustness.
\end{enumerate}

\section{Audience-specific Explanations}
We want to solve the problem of how to eliminate the target language audience’s incomprehension during the translation. With the help of the human translation solution, this problem can be transformed into another more specific problem, that is, how can we model audiences’ specific needs for additional information during translation? This means we need to develop a model that can predict which words will cause incomprehension to the target language audience during translation. In order to accurately predict the words that need to be explained when doing machine translation tasks, it is necessary to build a dataset for training and evaluation. In this paper we develop a methodology to identify these words in a parallel corpus. In a second step this data can then be used to train and test models that are able to predict these words.

Here is an example of translation with explanation for English$\to$German:
\begin{enumerate}
    \item \textbf{En}: \textbf{John Bunyan} said , “ He who runs from God in the morning will scarcely find Him the rest of the day . ”\\
    \textbf{De}: \textbf{John Bunyan} \textbf{\textit{, der Autor der bekannten Pilgerreise ,}} hat einmal gesagt : „ Wer morgens vor Gott wegläuft , wird Ihn den Rest des Tages kaum noch finden . “ 
\end{enumerate}

In German translation, \textbf{\textit{, der Autor der bekannten Pilgerreise ,}} is the explanation for \textbf{John Bunyan}, which tells the German audience that \textbf{John Bunyan} is a writer.

We started with an initial manual inspection of explanation. The main challenge thereby is that these explanations are extremely infrequent. On the other hand, the uncertainty in the position of the explanation also increases the difficulty of finding sentence pairs with explanations. To reduce the complexity of finding sentence pairs with explanations, we only consider the most common form of sentence pairs with explanations, that is, the explanation part immediately follows the word that needs to be explained. Based on our initial investigate we identified several key characteristics of sentences containing explanations. Based on these characteristics we build filters to find the sentences:
\begin{enumerate}
    \item The word being explained or the word in the phrase being explained is rare in the target language.
    \item The explanation is a redundant part of the sentence in the target language.
    \item The explanation follows the word or phrase being explained.
    \item The explanation contains punctuation.
    \item Words that differ from the word or phrase being explained are also included in the explanation.
    \item The word or phrase being explained is more likely to be a named entity.
    \item Information about words or phrases that need to be explained can be found using Wikipedia.
\end{enumerate}

%However, there are several problems when building the dataset. The first problem is which sentences contain words that require additional explanation. More precisely, how to distinguish sentences containing words that need to be explained from other sentences. If we can’t find a good way to distinguish the target sentences from other sentences, it will bring a huge manual workload to build the training dataset. In addition, in order for the trained model to have high quality, the training dataset must contain a sufficiently large number of target sentences. However, target sentences are not common, which makes how to find a sufficient number of target sentences become another problem that must be solved. Both of these problems require us to propose a sufficiently ideal method, which can find as many target sentences as possible while reducing the manual workload as much as possible. It also makes building a training dataset a difficult challenge.

\section{Identifying candidate source phrases}
Based on the summarized characteristics of sentence pairs with explanations, a heuristic method for efficiently searching sentence pairs with explanations is proposed. Considering the sparsity of the sentence pairs with explanations that need to be found, the goal of this method is to find as many sentence pairs with explanations as possible while minimizing the number of sentence pairs without explanations.

This heuristic method is divided into four processes. The first process is to identify words that may need explanation based on corpus statistics. The second process is to identify sentence pairs that may have explanation with the help of word alignment. After utilizing the internal knowledge we also integrated external knowledge. The third process is to use the named entity recognition (NER) model to identify the words that need to be explained. The last process is to exploit Wikipedia to more accurately identify target sentence pairs.

\subsection{Filtering based on corpus statistics}
Intuitively, when translating, if a word is rare in the target language, it is more likely to be explained than other words. In order to decide which words are rare, the word count within a certain range can be used. A word can be considered rare if its count is below a certain threshold. For the purpose of finding as many rare words as possible, the word count in all Wikipedia articles is used to check whether a word is rare. Meanwhile, if only the uncommon words in the target language are considered, it is found that many non-candidates are introduced in the experiment. So not only the word count in the target language but also the word count in the source language must be considered.

%However, if only the uncommon words in the target language are considered, it is found that many non-candidates are introduced in the experiment, so not only uncommon words in the target language but also uncommon words in the source language must be considered.

\subsection{Integrating word alignment inference}
While the model to identify the words that need to be explained does not have access to the translation, we can use the target side for identifying examples in the parallel data, in this case, this is even the most valuable way. With the help of the word alignment of the determined rare word and the word following it, the corresponding words in the target language sentence and their position in the sentence can be found. If there is a redundant part between the corresponding words in the target language sentence, it can be considered that there may be an explanation for the rare word in the redundant part.

%The next thing to determine is which sentence pairs may contain explanations. With the help of the word alignment of the determined rare word and the word following it, the corresponding words in the target language sentence and their position in the sentence can be found. If there is a redundant part between the corresponding words in the target language sentence, it can be considered that there may be an explanation for the rare word in the redundant part. What needs to be decided here is the length of the redundant part. If the length is too long, many possible examples will be missed, but if the length is too short, many non-candidates will be added.

%With the help of the word alignment of the determined rare word and the word following it, the corresponding words in the target language sentence and their position in the sentence can be found. If there is a redundant part between the corresponding words in the target language sentence, it can be considered that there may be an explanation for the rare word in the redundant part. What needs to be decided here is the length of the redundant part. If the length is too long, many examples will be missed, but if the length is too short, many non-candidates will be added.

The explanation part is often accompanied by punctuation marks, such as commas and parentheses. This means that it is possible to determine whether a redundant part contains an explanation by checking for possible punctuation in the redundant part. In addition, as the explanation contains additional information about the object being explained, the explanation should contain other words besides the explained word, so it can be judged whether there is an explanation in the redundant part by checking the words in the redundant part and their word alignment. If the redundant part also contains words other than the explained word, and none of the words in the redundant part have a word alignment, the redundant part can be considered as likely to contain the true explanation.

\begin{figure}[!ht]
\begin{center}
%\fbox{\parbox{6cm}{
%This is a figure with a caption.}}
% old picture \includegraphics[scale=0.5]{lrec2020W-image1.eps} 
\includegraphics[scale=0.25]{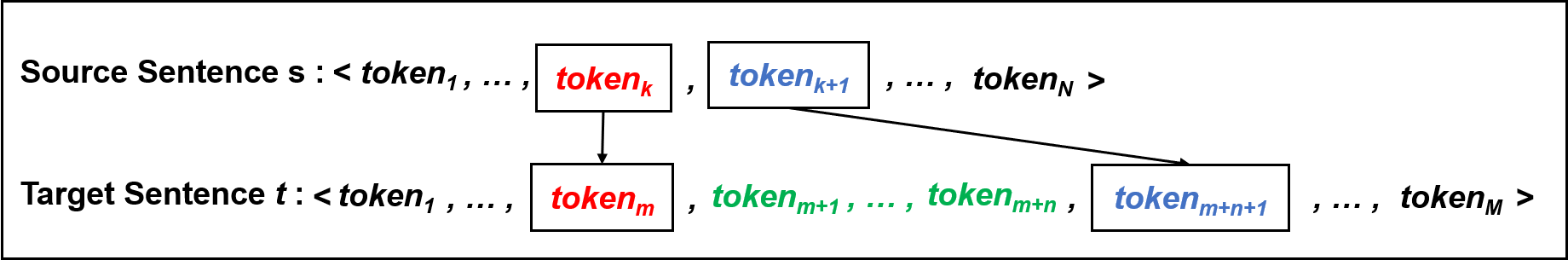} 
\caption{Candidate sentence pair}
\label{fig:sentence} 
\end{center}
\end{figure}

Depending on the form of the sentence pair that need to be identified and the summarized characteristics, we can visualize the candidate sentence pair with explanations. Figure \ref{fig:sentence} shows an ideal candidate sentence pair.

Given the source language sentence $\boldsymbol{s}$ and the corresponding target language translation sentence $\boldsymbol{t}$. The length of sentence $\boldsymbol{s}$ is $\boldsymbol{N}$, which means it is composed of $\boldsymbol{N}$ tokens. Similarly, the length of sentence $\boldsymbol{t}$ is $\boldsymbol{M}$. In sentence $\boldsymbol{s}$, the \emph{k-th} token $\boldsymbol{token_k}$ is an uncommon word, meanwhile, the \emph{m-th} token $\boldsymbol{token_m}$ aligned with $\boldsymbol{token_k}$ in sentence $\boldsymbol{t}$ is also a rare word. The next token $\boldsymbol{token_{k+1}}$ of token $\boldsymbol{token_k}$ in sentence $\boldsymbol{s}$ is aligned with token $\boldsymbol{token_{m+n+1}}$ in sentence $\boldsymbol{t}$. And in sentence $\boldsymbol{t}$ $\boldsymbol{token_{m+n+1}}$ is not the next token of $\boldsymbol{token_m}$, which means that there is a redundant part of length $\boldsymbol{n}$ after $\boldsymbol{token_m}$ in sentence $\boldsymbol{t}$. In the redundant part from $\boldsymbol{token_{m+1}}$ to $\boldsymbol{token_{m+n}}$, there are punctuation marks, such as $\boldsymbol{token_{m+1}}$ is likely to be a comma, or a parenthesis. All tokens from $\boldsymbol{token_{m+1}}$ to $\boldsymbol{token_{m+n}}$ in the redundant part should have no word alignment results, and should contain other words except $\boldsymbol{token_m}$.

\subsection{Using NER}
From the summarized characteristics about the target sentence pair we know that the word or phrase being explained is more likely to be a named entity. This provides another way to determine candidate sentence pairs. If the named entities in a sentence, such as person names, place names or organization names, can be identified and located, then the range of candidate sentences can be narrowed down better. Named entity recognition (NER) is an effective tool for identifying named entity in a sentence. NER can also give the location of each named entity. So NER can be used to further identify possible candidates while also reducing the number of non-candidates.

%The sixth characteristic summarized is that the words and phrases being explained are more likely to be proper nouns. This characteristic provides another way to determine candidate sentence pairs. If the proper nouns in a sentence, such as person names, place names or organization names, can be identified and located, then the range of candidate sentences can be narrowed down better. Named entity recognition (NER) is an effective tool for identifying proper nouns in a sentence. NER can identify the proper nouns that exist in a sentence, that is, named entities, and can also give the location of each named entity.

%So based on the sixth characteristic summarized, NER should further identify possible candidates while also reducing the number of non-candidates.

%After using NER to recognize all named entities, the words or phrases that need to be explained are likely to be in these named entities. Besides, the word being explained is either itself a named entity, or it is part of a named entity. So the candidate sentence pairs can be further determined by comparing the named entities with the previously confirmed words that may be explained.

The word being explained is either itself a named entity, or it is part of a named entity. So the candidate sentence pairs can be further determined by comparing the recognized named entities with the previously confirmed words that may be explained.

%On the other hand, we have also observed some examples as follows:
%\begin{enumerate}
%    \item \textbf{En}: Chevrolet also is the sole Engine supplier for the {\color{red}{Formula Rolon}} single seater series in India .\\
%    \textbf{De}: Chevrolet ist auch der alleinige Motorlieferant für die {\color{red}{Formel Rolon}} {\color{blue}{( Formel Rolon )}} einzelne seater Reihe in Indien .
%\end{enumerate}
%The redundant part of the sentence in the target language does not contain an explanation of the word or phrase, but the word or phrase itself. This problem can also be solved by using NER. After using NER, as long as the identified named entity is compared with the entity in the redundant part, these sentence pairs without explanations can be identified and removed.

%We also find another problem, that is, the redundant part of the sentence in the target language does not contain an explanation of the word or phrase, but the word or phrase itself. This problem can also be easily solved by using NER. After using NER, as long as the identified named entity is compared with the entity in the redundant part, the sentence pair that contains only the named entity itself in the redundant part can be identified.

\subsection{Using Wikipedia}
All the named entities in a sentence can be recognized after NER. Based on named entities, another method that can further to identify target sentence pairs with explanations is using Wikipedia. This step is also inspired by the work of \cite{nothman2008transforming,nothman2009analysing,nothman2013learning}. Their work confirms that Wikipedia can be used to build dataset for training NER models  and NER models can also achieve optimal results.

For most phrases that need to be explained, the corresponding articles can be searched in Wikipedia, so the titles of Wikipedia articles can be used to determine target sentence pairs. If a source language named entity is a title of a Wikipedia article, then it is likely a candidate that needs to be explained. However, this only considers the aspect of the source language, if the consideration for the target language is added, candidates can be further identified. If a source language named entity is the title of a Wikipedia article, and the corresponding target language named entity is not the title of a Wikipedia article, this means that the source language audience has a source of information to understand this named entity, while the target language audience does not have a corresponding information source. In this case, this named entity is more likely to be a good candidate that needs to be explained.

\begin{figure}[!ht]
\begin{center}
%\fbox{\parbox{6cm}{
%This is a figure with a caption.}}
% old picture \includegraphics[scale=0.5]{lrec2020W-image1.eps} 
\includegraphics[scale=0.25]{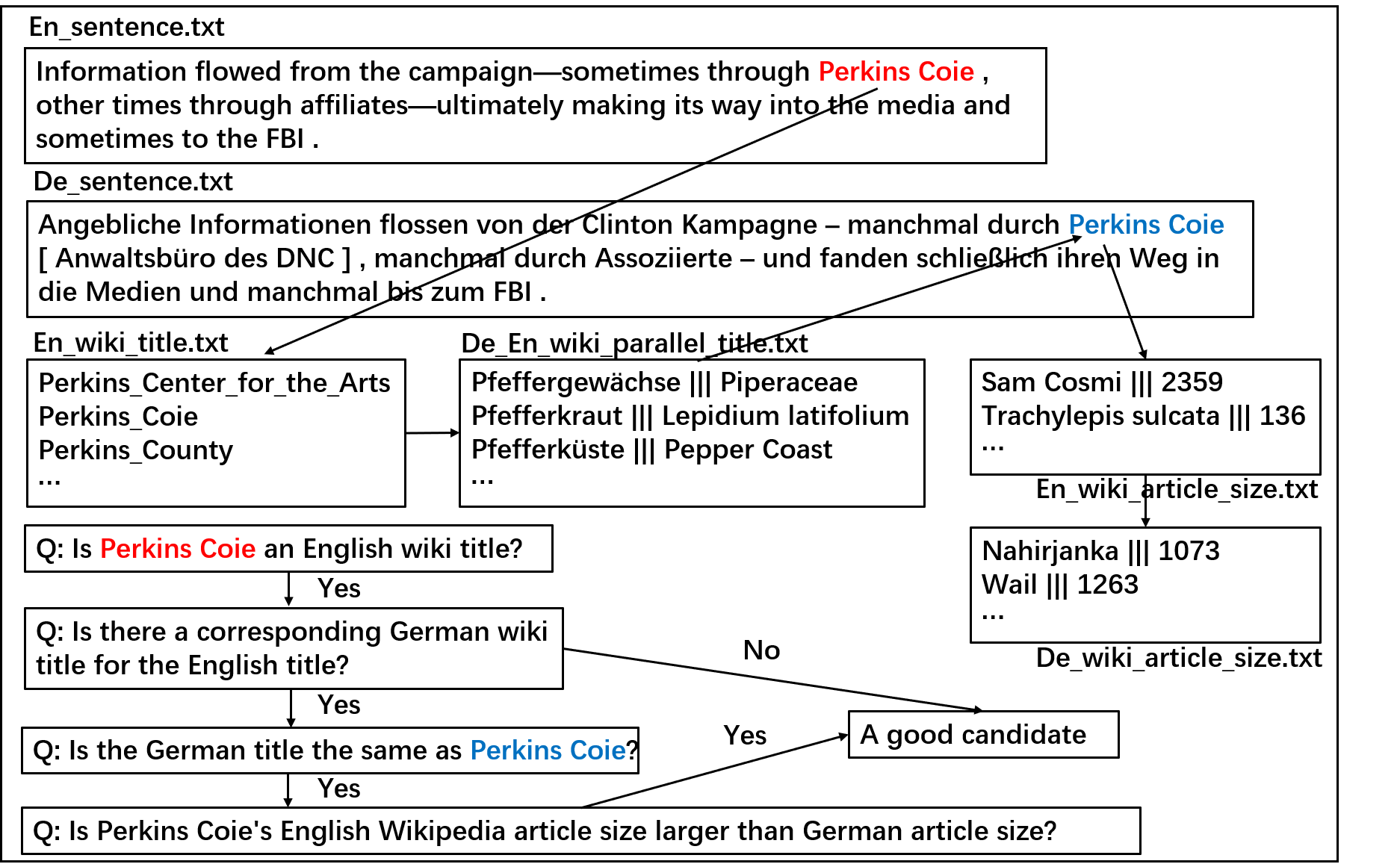} 
\caption{An example of identifying candidates using Wikipedia}
\label{fig:wiki_iden} 
\end{center}
\end{figure}

On the other hand, in addition to the title of the Wikipedia article can be used to determine candidates, the Wikipedia article itself can also be used to determine candidates. If both the named entity in the source language and the corresponding named entity in the target language are the titles of Wikipedia articles, then the articles corresponding to the titles can be compared. More precisely, candidates can be determined by comparing the size of Wikipedia articles. If the size of the Wikipedia article in the source language is larger than the size of Wikipedia article in the target language, this situation means that while both the source and target language audiences have a source of information to understand the candidate, but the candidate is less common in target language, then the title of the source language Wikipedia article might be a good candidate that needs to be explained. Figure \ref{fig:wiki_iden} shows the process of using Wikipedia to identify candidates.

There are many text comparisons in this step, for example, the comparisons between named entities and Wikipedia titles. The possible form inconsistencies between them will make the comparison difficult. In order to simplify the text comparison when using Wikipedia, the stemming algorithm will be applied.

\section{Evaluation}
\subsection{Setup}
We choose CCMatrix \citep{schwenk2019ccmatrix,fan2021beyond} as the corpus required for the experiment. All CCMatrix data are downloaded from OPUS \citep{tiedemann2012parallel}. The version of Wikipedia database backup dumps used in the experiments is 20221101. We use wikiextractor \footnote{https://github.com/attardi/wikiextractor} to extract Wikipedia articles. Meanwhile, we use the tool wikipedia-parallel-titles \footnote{https://github.com/clab/wikipedia-parallel-titles} to create the Wikipedia parallel titles corpus. For preprocessing, we choose spaCy \citep{Honnibal_spaCy_Industrial-strength_Natural_2020} as the word tokenization tool. For Chinese word tokenization, we use pkuseg \citep{pkuseg} under the framework of spaCy. We use awesome-align \citep{dou2021word} to extract word alignment results. We choose the Snowball algorithm from NLTK \citep{bird2009natural} to perform stemming to simplify text comparison. When using word alignment to find the target sentence pair, the length of the redundant part in the target language sentence is set to be greater than or equal to 3.

\subsection{Evaluation Metric}
Filter the target sentences from the corpus to construct the training dataset is a classification problem. Therefore, the metric BLEU does not work here. For a classification problem, F1-score is a good evaluation metric. The calculation of the F1-score requires the number of positive examples and the number of negative examples. However, due to the sparsity of sentences with explanations, we only focus on whether the target sentence can be found, which means that in our experiments, the evaluation of the method for finding target sentences only involves sentences with explanation, i.e. only the number of positive examples is considered. This means that we just selected a subset of F1-score as the evaluation metric for our experiments.

%F1-score is selected and used as the evaluation metric. The calculation of the F1-score requires the number of positive examples and the number of negative examples. However, in our experiments, the evaluation only involves target sentences, i.e. only the number of positive examples is considered. This means that we just selected a subset of F1-score as the evaluation metric for our experiments.

\subsection{Initial Result}
We first focus only on the English$\to$German (En$\to$De) language pair. In order to compare and evaluate the subsequent steps of our proposed method, we first run our method to the step before NER, i.e., using only word counts and word alignments to find target sentence pair candidates. The first 5 million sentence pairs from the corpus are taken as the input, and the word count thresholds for both source and target languages are set to 15000. This means that if a word, its word count is below 15000, it is considered a rare word. The results are in Table \ref{tab:5_mil_init_res}.

\begin{table}[!ht]
    \centering
    \begin{tabular}{|l|l|l|}
         \hline
         \textbf{En$\to$De} & \textbf{Num. of sentences } & \textbf{F1}\\
         \hline
         Total & 5000000 & -\\
         \hline
         1. Initial result & 8977/173 & $3.78\%$\\
         \hline
    \end{tabular}
    \caption{The initial results of the En$\to$De}
    \label{tab:5_mil_init_res}
\end{table}

It can be found from the Table \ref{tab:5_mil_init_res} that for En$\to$De there are still 8977 sentence pairs remaining. On the basis of the remaining sentence pairs, manual work is performed to select out the sentence pairs that contain explanations. Finally, 173 sentence pairs with explanations are found out of 8977 remaining sentence pairs. This means that only 1.92$\%$ of the original sentences contain explanations. F1-score is also calculated and given in the table. 

%The same manual work is done for the input sentence pairs of 5 million to select the sentence pairs containing the explanation. The statistical results are shown in Table \ref{tab:5_mil_man_work_stat_res}. The corresponding F1-score for each language pair is also calculated. Based on the results in Table \ref{tab:5_mil_man_work_stat_res},  the subsequent experiments can be performed to compare the effects of different NER tools.

\subsection{Follow-up results}
Based on the results in Table \ref{tab:5_mil_init_res}, the subsequent experiments can be performed. Since the models provided by different NER tools and different word count thresholds will affect the final experimental results, different combinations of NER models and thresholds need to be considered and compared to obtain the best results. We compared the performance of the NER models from Flair \citep{akbik2019flair}, spaCy \citep{Honnibal_spaCy_Industrial-strength_Natural_2020}, and Stanza \citep{qi2020stanza}. We also tried five different word count thresholds for both source and target languages under each NER model: 100, 1000, 5000, 10000 and 15000. We found that for En$\to$De, optimal results can be obtained when using the NER model provided by Flair with a word count threshold of 5000 (i.e., the word count is under 5000) for both the source and target languages. The final overall results are in Table \ref{tab:5_mil_res_all}. 

%We found that for En$\to$De, optimal results can be obtained when using the NER model provided by Flair (\cite{akbik2019flair}) with a word count threshold of 5000 for both the source and target languages. The final overall results are in Table \ref{tab:5_mil_res_all}. 

%We found that for En$\to$De, optimal results can be obtained when using the NER model provided by Flair (\cite{akbik2019flair}) with a word count threshold of 5000 for both the source and target languages, i.e., the word count is under 5000. The final overall results are in Table \ref{tab:5_mil_res_all}. 

\begin{table}[!ht]
    \centering
    \begin{tabular}{|l|l|l|}
         \hline
         \textbf{Step} & \textbf{Num. of sent.}& \textbf{F1}\\
         \hline
         Initial result & 8977/173 & $3.78\%$\\
         \hline
         \makecell{1. Using word count\\ and alignment} & 3102/134 & $8.18\%$\\
         \hline
         2. Using NER & 791/93 & $19.29\%$\\
         \hline
         3. Using Wiki & 323/44 & $17.74\%$\\
         \hline
    \end{tabular}
    \caption{The final result of the En$\to$De}
    \label{tab:5_mil_res_all}
\end{table}

Comparing the initial result and the result of the first step, we can find that as the word count threshold becomes smaller, the number of remaining sentence pairs decreases, and the number of target sentence pairs with explanations in the remaining sentence pairs also decreases. But on the other hand, the F1-score increased from $3.78\%$ to $8.18\%$. Comparing the results of the first step and the second step, we can find that after using NER, the number of remaining sentence pairs and target sentence pairs both decreased, but the F1-score increased from $8.18\%$ to $19.29\%$. The significant improvement in the F1-score shows that NER is a powerful and effective tool for identifying target sentence pairs. 

Finally, comparing the results of the second step with the third step, we can find that after using Wikipedia to continue to identify target sentence pairs, the number of remaining sentence pairs and target sentence pairs are reduced. Meanwhile, the F1-score is also decreased from $19.29\%$ to $17.74\%$. This shows that Wikipedia can greatly reduce manual work while losing a relatively small F1-score.

%Finally, comparing the results of the second step with the third step, we can find that after using Wikipedia to continue to identify target sentence pairs, the number of remaining sentence pairs and target sentence pairs are reduced. Meanwhile, the F1-score is also decreased from $19.29\%$ to $17.74\%$. However, combined with the results of the number of sentence pairs and the F1-score, it can be considered that the use of Wikipedia can greatly reduce manual work and at the same time an acceptable F1-score is also obtained. This shows that Wikipedia is also a helpful tool for identifying target sentence pairs.

The results in Table \ref{tab:5_mil_res_all} prove that our proposed method can greatly improve the efficiency of finding target sentence pairs. In order to verify the general effectiveness of our method, we also test the effect of our method on other inputs. We randomly select 5 million sentence pairs from all remaining sentence pairs in the corpus except for the first 5 million sentence pairs. These randomly selected sentence pairs are used as input. The same configurations are also used for test. In order to avoid accidental errors, we conduct five experiments for testing. The results of the five experiments for testing are very similar, so we only provide the results of one of them. The results of the experiments for testing are in Table \ref{tab:en_de_test_res} and Table \ref{tab:en_de_percentage_res}.

\begin{table}[!ht]
    \centering
    \begin{tabular}{|l|l|l|l|l|l|l|}
         \hline
         \textbf{Step} & \textbf{Num. of sent.}\\
         \hline
         \makecell{1. Using word count \& align.} & 17271\\
         \hline
         2. Using NER & 7176\\
         \hline
         3. Using Wiki & 2832\\
         \hline
    \end{tabular}
    \caption{The results of the experiment for testing (En$\to$De)}
    \label{tab:en_de_test_res}
\end{table}

\begin{table}[!ht]
    \centering
    \begin{tabular}{|l|l|l|}
         \hline
         \textbf{En$\to$De} & \textbf{Num. of sent.} & \textbf{Perc.}\\
         \hline
         3. Wiki (First 5M) & 323/44 & 13.62$\%$\\
         \hline
         3. Wiki (Rand. 5M) & 2832/294 & 10.38$\%$\\
         \hline
    \end{tabular}
    \caption{The percentage results of En$\to$De}
    \label{tab:en_de_percentage_res}
\end{table}

If we compare the results in Table \ref{tab:en_de_test_res} with those in Table \ref{tab:5_mil_res_all}, we can find a non-negligible gap between the two results regarding the number of remaining sentence pairs. In order to check the proportion of target sentence pairs with explanations in the remaining sentence pairs, we also check the number of target sentence pairs among the last remaining sentence pairs. We select the results of the last step (i.e. the step to use Wikipedia) for validation. The proportion results are in Table \ref{tab:en_de_percentage_res}. 

When the input is the first 5 million sentence pairs of the corpus, 44 sentence pairs with explanations can be found in the remaining 323 sentence pairs. When 5 million randomly selected sentence pairs are taken as input, we found 294 target sentence pairs with explanations out of 2832 remaining sentence pairs. This means that when the input is random, the number of remaining sentence pairs and the sentence pairs with explanations are all increased. However, when we check the proportion results, we can find that for the random input, among the remaining sentence pairs we can find $10.38\%$ target sentence pairs. And for the first 5 million sentence pairs, the proportion is $13.62\%$. Although the proportion result of the random input is lower than that of the first 5 million sentence pairs, it is still higher than $10\%$, which is an acceptable result.

The results in Table \ref{tab:en_de_percentage_res} illustrate that our proposed method for finding sentence pairs with explanations is robust against different input data and efficient for the En$\to$De. Our proposed method is independent of the distribution of data, and a large number of non-target sentence pairs can be removed. Therefore, the number of last remaining sentence pairs is extremely small. Among the last remaining sentences, no matter how the number of remaining sentences changes, more than $10\%$ of the target sentence pairs can always be found.

%Finally, we also check whether each named entity that is explained in the found target sentence pairs also always needs to be explained in other sentences. Based on 44 named entities that require explanation found in experiments, for each named entity found, all sentence pairs containing this named entity will be found in the input sentence pairs, each sentence pair is then checked to see if it contains an explanation for the named entity. The result is in Figure \ref{fig:en_de_ne_prop_dis}.

\begin{figure}[!ht]
\begin{center}
%\fbox{\parbox{6cm}{
%This is a figure with a caption.}}
% old picture \includegraphics[scale=0.6]{lrec2020W-image1.eps} 
\includegraphics[scale=0.5]{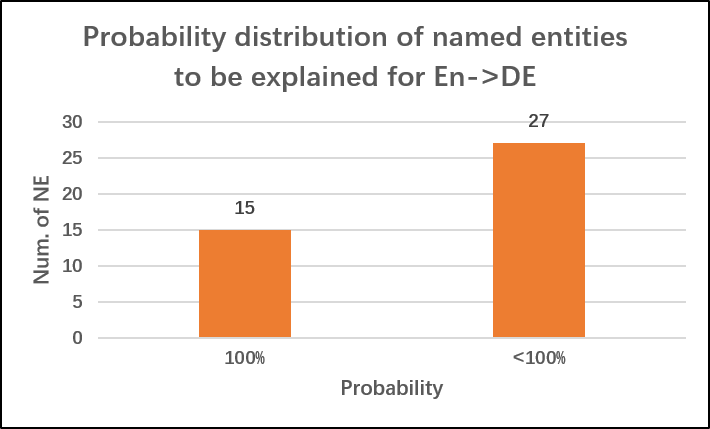} 
\caption{Probability distribution of NE to be explained}
\label{Fig.Probability_distribution} 
\end{center}
\end{figure}

\begin{figure}[!ht]
\begin{center}
%\fbox{\parbox{6cm}{
%This is a figure with a caption.}}
% old picture \includegraphics[scale=0.6]{lrec2020W-image1.eps} 
\includegraphics[scale=0.5]{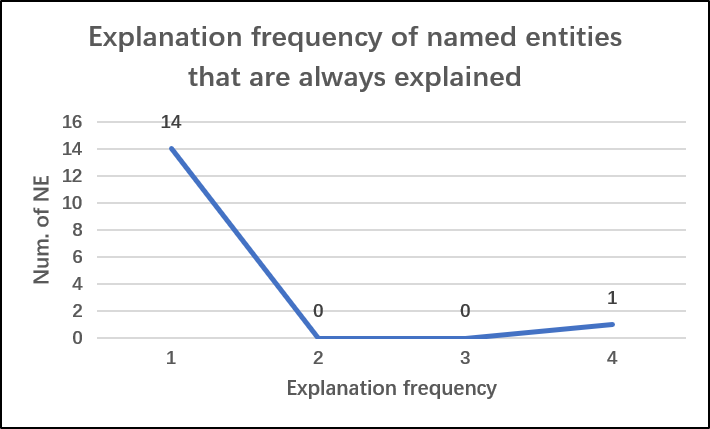} 
\caption{Explanation frequency of NE that is always explained}
\label{Fig.Number_times} 
\end{center}
\end{figure}

Finally, we also check whether each named entity that is explained in the found target sentence pairs also always needs to be explained in other sentences. Based on 44 named entities that require explanation found in experiments, in the input of 5 million sentence pairs, for each named entity, we check whether each sentence pair containing this named entity also contains an explanation for it. The result is in Figure \ref{Fig.Probability_distribution} and Figure \ref{Fig.Number_times}.

%\begin{figure}[htbp]
%\begin{center}
%  \includegraphics[width = 0.5\textwidth]{figures/EN-DE.png}
%  \caption{\label{fig:en_de_ne_prop_dis}Probability distribution of named entities to be explained for En$\to$De}
%\end{center}
%\end{figure}

%\begin{figure}[htbp]
%\begin{center}
%  \includegraphics[width = 0.5\textwidth]{figures/fig_num_times_EN-DE.png}
%  \caption{\label{fig:en_de_ne_num_times}Probability distribution of named entities to be explained for En$\to$De}
%\end{center}
%\end{figure}

%Figure \ref{Fig.main} has two sub figures, fig. \ref{Fig.sub.1} is the travel demand of driving auto, and fig. \ref{Fig.sub.2} is the travel demand of park-and-ride.

After removing duplicate named entities, there are 42 explained named entities left for En$\to$De. From the Figure \ref{Fig.Probability_distribution} we can find that not all named entities always need to be explained. Only 15 named entities are always explained (with 100$\%$ probability). Based on these 15 named entities that always require explanation, we also examine how often these named entities are explained, the result is in the Figure \ref{Fig.Number_times}. Among the 15 named entities that are always explained, 14 named entities are explained only once, and only 1 named entity is explained 4 times.

\subsection{Multi-language results}
We also conduct experiments for English$\to$French (En$\to$Fr) and English$\to$Chinese (En$\to$Zh) language pairs. The first 5 million sentence pairs in the corpus are still used as the input. For initial results, the word count thresholds for both the source and target languages are set to 15000. We found that for En$\to$Fr, optimal results can be obtained when using the NER model provided by Stanza \citep{qi2020stanza} with a word count threshold of 5000 for both the source and target languages. And for En$\to$Zh, optimal results can be obtained when using the NER model provided by HanLP \citep{he-choi-2021-stem} with a word count threshold of 5000 for both the source and target languages. Although the word count threshold to obtain the optimal results is the same for all language pairs. However, for each language pair, the results of the number of remaining sentence pairs and the number of sentence pairs with explanations are different.

Table \ref{tab:5_mil_en_fr_compelete_res} is the result for En$\to$Fr. The result for En$\to$Zh is in Table \ref{tab:5_mil_en_zh_compelete_res}. The results for these two language pairs are similar to those for En$\to$De. As the word count threshold gets smaller, the number of remaining and target sentence pairs decreases, while the F1-score rises. NER is also a powerful and effective tool for identifying target sentence pairs for both language pairs. After using NER, although the number of remaining sentence pairs and target sentence pairs are further reduced, the F1-score is significantly improved. And the use of Wikipedia can still achieve an acceptable F1-score while greatly reducing manual work.

\begin{table}[!ht]
    \centering
    \begin{tabular}{|l|l|l|}
         \hline
         \textbf{Step} & \textbf{Num. of sent.} & \textbf{F1}\\
         \hline
         Initial Result & 6982/122 & $3.43\%$\\
         \hline
         \makecell{1. Using word count\\ and alignment} & 3350/95 & $5.47\%$\\
         \hline
         2. Using NER & 1332/76 & $10.45\%$\\
         \hline
         3. Using Wiki & 395/18 & $6.96\%$\\
         \hline
    \end{tabular}
    \caption{The final results of En$\to$Fr language pair}
    \label{tab:5_mil_en_fr_compelete_res}
\end{table}

\begin{table}[!ht]
    \centering
    \begin{tabular}{|l|l|l|}
         \hline
         \textbf{Step} & \textbf{Num. of sent.}& \textbf{F1}\\
         \hline
         Initial Result &13541/402 & $5.77\%$ \\
         \hline
         \makecell{1. Using word count\\ and alignment} & 7360/302 & $7.78\%$\\
         \hline
         2. Using NER & 2557/194 & $13.11\%$\\
         \hline
         3. Using Wiki & 1083/87 & $11.72\%$\\
         \hline
    \end{tabular}
    \caption{The final results of En$\to$Zh language pair}
    \label{tab:5_mil_en_zh_compelete_res}
\end{table}

%Similarly, we also test the effect of our method on random inputs for En$\to$Fr and En$\to$Zh language pairs. The results are in Table \ref{tab:en_fr_test_res} and Table \ref{tab:en_zh_test_res}. The results for En$\to$Fr and En$\to$Zh are similar to those for En$\to$De. We can find that the results of the five experiments for testing are very similar. Besides, compared to the experimental results of the first five million sentence pairs as input, when the input is random, a non-negligible gap regarding the number of remaining sentence pairs can be always found.

Similarly, we also test the effect of our method on random inputs for En$\to$Fr and En$\to$Zh language pairs. The results are in Table \ref{tab:en_fr_test_res} and Table \ref{tab:en_zh_test_res}. In order to avoid accidental errors, we also conduct five experiments. The results of the five experiments for En$\to$Fr and En$\to$Zh are also very similar, so we also provide the results of one of them. Compared to the experimental results of the first five million sentence pairs as input, when the input are random five million sentence pairs, a non-negligible gap regarding the number of remaining sentence pairs can be always found.

\begin{table}[!ht]
    \centering
    \begin{tabular}{|l|l|l|l|l|l|l|}
         \hline
         \textbf{Step} & \textbf{Num. of sent.}\\
         \hline
         1. Using word count \& align. & 22679\\
         \hline
         2. Using NER & 12420\\
         \hline
         3. Using Wiki & 4051\\
         \hline
    \end{tabular}
    \caption{The results of the experiments for testing (En$\to$Fr)}
    \label{tab:en_fr_test_res}
\end{table}

\begin{table}[!ht]
    \centering
    \begin{tabular}{|l|l|l|l|l|l|l|}
         \hline
         \textbf{Step} & \textbf{Num. of sent.}\\
         \hline
         1. Using word count \& align. & 16267\\
         \hline
         2. Using NER & 7451\\
         \hline
         3. Using Wiki & 3149\\
         \hline
    \end{tabular}
    \caption{The results of the experiments for testing (En$\to$Zh)}
    \label{tab:en_zh_test_res}
\end{table}

We also calculated the proportion of target sentence pairs with explanations in the results of experiments for validation for En$\to$Fr and En$\to$Zh. The proportion results are in Table \ref{tab:percentage_res_en_fr} and Table \ref{tab:percentage_res_en_zh}. For En$\to$Fr, a surprising result is obtained. For the results with the random input, $8.24\%$ of the target sentence pairs can be found in the last remaining 4051 sentence pairs, which is much higher than the $4.56\%$ of the results with the first 5 million sentence pairs as the input. Due to the identical experimental parameters, the significant difference in proportion results is presumably caused by the distribution of input data. For En$\to$Zh, $7.40\%$ of the target sentence pairs can be found in the remaining 3149 sentence pairs when the input is random, which is very close to the $8.03\%$ of the results when the input is the first 5 million sentence pairs. This means that our proposed method is also robust against different input data and efficient for En$\to$Fr and En$\to$Zh. The number of last remaining sentence pairs is extremely small. For En$\to$Fr, more than $5\%$ of the target sentence pairs can always be found among the last remaining sentence pairs. While for En$\to$Zh, more than $7\%$ of the target sentence pairs can always be found in the last remaining sentence pairs.

\begin{table}[!ht]
    \centering
    \begin{tabular}{|l|l|l|}
         \hline
         \textbf{En$\to$Fr} & \textbf{Num. of sent.} & \textbf{Perc.}\\
         \hline
         3. Wiki (First 5M.) & 395/18 & 4.56$\%$\\
         \hline
         3. Wiki (Rand. 5M.) & 4051/334& 8.24$\%$\\
         \hline
    \end{tabular}
    \caption{The percentage results of En$\to$Fr}
    \label{tab:percentage_res_en_fr}
\end{table}

\begin{table}[!ht]
    \centering
    \begin{tabular}{|l|l|l|}
         \hline
         \textbf{En$\to$Zh} & \textbf{Num. of sent.} & \textbf{Perc.}\\
         \hline
         3. Wiki (First 5M.) & 1083/87 & 8.03$\%$\\
         \hline
         3. Wiki (Rand. 5M.) & 3149/233& 7.40$\%$\\
         \hline
    \end{tabular}
    \caption{The percentage results of En$\to$Zh}
    \label{tab:percentage_res_en_zh}
\end{table}

We can also provide some examples with explanations found for En$\to$Fr and En$\to$Zh.
\begin{enumerate}
    \item \textbf{En}: ‘ Even if \textbf{WMO} agrees , I will still not pass on the data .\\
    \textbf{Fr}: « Même si la \textbf{WMO} \textbf{\textit{[ Organisation mondiale de la météorologie ]}} est d’ accord , je ne vais toujours pas transmettre les données .
    
    \item \textbf{En}: I looked at \textbf{Raytheon} ’s annual report and they lost a lot of money in this area , but they ’re selling their own planes so maybe they do n’t care .\\
    \begin{CJK}{UTF8}{gbsn}
    \textbf{Zh}: 我 看 了 \textbf{Raytheon} \textbf{\textit{[ 美国 主要 的 军事 承包 商]}} 的 年报 ， 它们 在 这个 领域 亏 了 很多 钱 ， 但 他们 销售 自己 的 飞机 ， 因此 可能 并不 在意 。
    \end{CJK}
\end{enumerate}

%Finally, for En$\to$Fr and En$\to$Zh language pairs, we also check whether each named entity that is explained in the found target sentence pairs also always needs to be explained in other sentences. The result is in Figure \ref{fig:en_fr_ne_prop_dis} and Figure \ref{fig:en_zh_ne_prop_dis}. From these figures we can find that for En$\to$Fr and En$\to$Zh, not all named entities also always need to be explained. For both language pairs, there are some named entities that need to be explained with probability lower than $10\%$.

%\begin{figure}[htbp]
%\begin{center}
%  \includegraphics[width = 0.6\textwidth]{figures/EN-FR.png}
%\caption{\label{fig:en_fr_ne_prop_dis}Probability distribution of named entities for En$\to$FR}
%\end{center}
%\end{figure}

%\begin{figure}[htbp]
%\begin{center}
%  \includegraphics[width = 0.6\textwidth]{figures/EN-ZH.png}
%\caption{\label{fig:en_zh_ne_prop_dis}Probability distribution of named entities for En$\to$Zh}
%\end{center}
%\end{figure}

\section{Conclusion}
We propose a heuristic method to find target sentence pairs with explanations. In this method, both internal and external knowledge are utilized: word count, word alignment, named entity recognition, and Wikipedia. We conduct experiments on three language pairs: English$\to$German, English$\to$French and English$\to$Chinese. The results show that for each language pair, our proposed method can reduce the number of remaining sentence pairs to an extremely low number. Moreover, our method is robust, among the remaining sentence pairs, a certain proportion of target sentence pairs can always be found for each language pair. Among the remaining sentence pairs, more than $10\%$ of the target sentence pairs can be found for the English$\to$German, more than $7\%$ of the target sentence pairs can be found for the English$\to$Chinese, and for the English$\to$French, more than $5\%$ of the target sentence pairs can be found. This means that a sufficient number of target sentence pairs can be efficiently found using our method so that we can construct a training dataset for the training of the model in the future.

\nocite{*}
\section{Bibliographical References}\label{sec:reference}

\bibliographystyle{lrec-coling2024-natbib}
\bibliography{lrec-coling2024-example}

%\section{Language Resource References}
%\label{lr:ref}
%\bibliographystylelanguageresource{lrec-coling2024-natbib}
%\bibliographylanguageresource{languageresource}

\end{document}